\documentclass{ecai} 

\usepackage{latexsym}
\usepackage{amssymb}
\usepackage{amsmath}
\usepackage{amsthm}
\usepackage{booktabs}
\usepackage{enumitem}
\usepackage{graphicx}
\usepackage{color}

\usepackage[ruled]{algorithm2e} 
\usepackage{cleveref} 
\usepackage{multirow} 
\usepackage{booktabs} 
\usepackage{float} 
\usepackage{subfig}
\usepackage[cmyk]{xcolor} 

\def\name{\textit{FedReMa}\xspace}
\newcommand{\BibTeX}{B\kern-.05em{\sc i\kern-.025em b}\kern-.08em\TeX}
\newcommand{\bm}[1]{\boldsymbol{#1}}

\newcommand{\tbf}[1]{\textbf{#1}}

\begin{document}


\begin{frontmatter}


\paperid{123} 


\title{FedReMa: Improving Personalized Federated Learning via Leveraging the Most Relevant Clients}



\author[A]{\fnms{Han}~\snm{Liang}}
\author[A]{\fnms{Ziwei}~\snm{Zhan}}
\author[A]{\fnms{Weijie}~\snm{Liu}} 
\author[A]{\fnms{Xiaoxi}~\snm{Zhang}\thanks{Corresponding author: Xiaoxi Zhang (zhangxx89@mail.sysu.edu.cn)}}
\author[B]{\fnms{Chee Wei}~\snm{Tan}} 
\author[A]{\fnms{Xu}~\snm{Chen}} 

\address[A]{Sun Yat-sen University}
\address[B]{Nanyang Technological University}


\begin{abstract}
Federated Learning (FL) is a distributed machine learning paradigm that achieves a globally robust model through decentralized computation and periodic model synthesis, primarily focusing on the global model's accuracy over aggregated datasets of all participating clients. Personalized Federated Learning (PFL) instead tailors exclusive models for each client, aiming to enhance the accuracy of clients' individual models on specific local data distributions. Despite of their wide adoption, existing FL and PFL works have yet to comprehensively address the class-imbalance issue, one of the most critical challenges within the realm of data heterogeneity in PFL and FL research. In this paper, we propose \name, an efficient PFL algorithm that can tackle class-imbalance by 1) utilizing an {\em adaptive inter-client co-learning} approach to identify and harness different clients' expertise on different data classes throughout various phases of the training process, and 2) employing distinct aggregation methods for clients' feature extractors and classifiers, with the choices informed by the different roles and implications of these model components. Specifically, driven by our experimental findings on inter-client similarity dynamics, we develop {\em critical co-learning period} (CCP), wherein we introduce a module named {\em maximum difference segmentation} (MDS) to assess and manage task relevance by analyzing the similarities between clients' logits of their classifiers. Outside the CCP, we employ an additional scheme for model aggregation that utilizes historical records of each client's most relevant peers to further enhance the personalization stability. We demonstrate the superiority of our \name in extensive experiments. The code is available at \url{https://github.com/liangh68/FedReMa}.

\end{abstract}

\end{frontmatter}

\section{Introduction}

Federated learning (FL) enables collaborative model training across decentralized devices or data sources, preserving privacy and scalability while harnessing the collective intelligence of distributed data~\citep{fedavg}. However, this algorithm based on periodic synchronization requires all client models to share the same structure and parameters, which brings about huge heterogeneity problems~\citep{zhao2018noniid,zhu2021noniidsurvey,zhang2024coded}.

Unlike traditional FL approaches that view data heterogeneity across clients as an obstacle, personalized federated learning (PFL) aims to exploit these data variations and train distinct models so as to tailor clients' individual models to their local optimization subproblems under heterogeneous data distributions, focusing more on model personalization. 
This paradigm shift in machine learning opens up new possibilities for personalized services, recommendations, and predictions, all while respecting user privacy.

Nevertheless, PFL still encounters numerous challenges. Firstly, disparate data distributions arise from varying data preferences, causing label skew and feature skew, known as the non-IID data challenge. Among all the subcategories of non-IID data issue, class-imbalance is arguably the most challenging one. While numerous studies have dedicated to solving class-imbalance in centralized machine learning settings~\citep{wang2018towards,rethinking,learningnot}, these methods are not readily applicable to PFL, given its unique challenges posed by the distributed and privacy-preserving nature, which prevents sharing individual client data to balance the overall dataset. Moreover, uniform aggregation operations adopted in most FL studies employed during each period of model synchronization can be detrimental from a local perspective~\citep{fedrod}. 
This occurs because each model, meticulously trained for a specific client's data, may amalgamate with models harboring {\em unfavorable knowledge} from diverse distributions of other clients. Therefore, numerous PFL studies propose various methods to mitigate performance degradation due to standard model aggregation under non-IID data. Many of the leading PFL studies focus on retaining personalization layers locally~\citep{fedbabu,fedper,fedrep,fedrod,lgfedavg} or adapting them to local tasks through certain modifications~\citep{fedproto,CReFF,fedclassavg,pfedme,perfedavg,liu2024decentralized, tamirisa2024fedselect}. 

\begin{figure}[t]
\centering
\includegraphics[width=.45\textwidth]{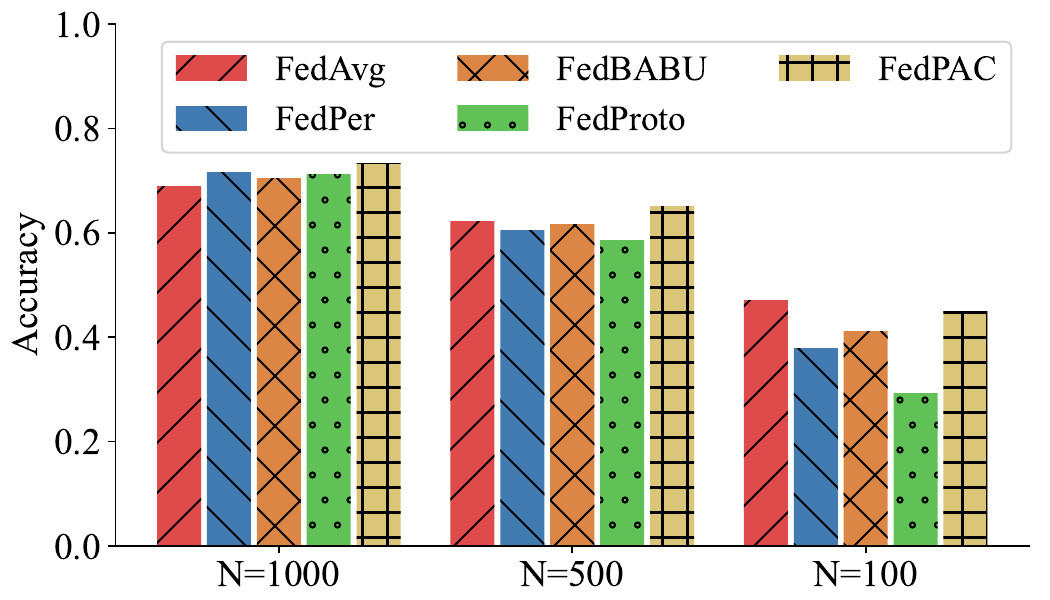}
\caption{Average local accuracy of advanced PFL methods can even fall below FedAvg when the number of trainable samples decreases.}
\label{fig:motivation}
\vskip 0.5cm
\end{figure}

However, in practice, PFL may encounter scenarios where clients' data can be both sparse and imbalanced. Most of the PFL works do not fully leverage the {\em dynamic} inter-client relationships, which should be exploited to improve model generalization along with personalization~\citep{fedbabu,fedper,fedrep,pfedme,fedfa}.
Moreover, methods using personalized layers, represented by FedPer~\citep{fedper} and FedRep~\citep{fedrep}, do not enable {\em classifier synchronization} among clients and the PS, and they show disappointing performance.
where an extensive network encompasses a multitude of clients with sparse data, PFL methods struggle significantly with generalization. Our experiments (Figure~\ref{fig:motivation}) demonstrate that, even the vanilla algorithm FedAvg~\citep{fedavg} can outperform PFL methods when the number data samples decreases. Besides, the method like FedPer, which disables classifier sharing, suffer from a more obvious performance degradation than FedPAC~\cite{fedpac}, a representative method that leverages classifier collaboration. This result reveals the importance of the knowledge transfer from clients' classifiers for improving generalization in PFL. We then gain our first insight: {\em generalization capability needs to be retained for PFL, even though personalization is its primary goal}. 

{\bf Exploiting inter-client classifier collaboration.} Given the above observation, it is imperative to employ a method that can gauge the underlying correlation and leverage the valuable inter-client knowledge to optimize local classifiers. Some studies~\citep{cho2020client,goetz2019active} suggest assigning higher selection probabilities to clients with larger loss values for aggregation. However, selection based on training indicators often ignores the correlation between clients. A few works improve this strategy. 
FedPer++~\citep{fedper++} uses a feature regularization training strategy to mitigate local overfitting risks and reduce parameter differences between clients, thereby promoting global feature extractor aggregation. A more recent work, FedPAC~\citep{fedpac}, quantifies the benefit of a combination of classifiers for each client as a function of the combination weights and derives an optimization problem to estimate the optimal weights. 
Yet, acquiring the optimal weights in FedPAC necessitates solving a complex optimization problem, significantly prolonging training time. Moreover, this algorithm still exhibits suboptimal performance in scenarios involving sparse or homogeneous data. Therefore, existing classifier collaboration needs to be improved in terms both time efficiency and the ability to boost model generalizability and personalizability. Intuitively, these two capabilities might seem to conflict with each other. However, our second insight implies that, {\em selectively knowledge utilization from peers, whose models can preserve and enhance the strength in its own dominant classes while improving performance in other classes without compromising their existing strength, can improve both generalization and personalization.} 
An intrinsic question then arises: How to design a lightweight method to identify the most relevant peers and effectively utilize their knowledge for each individual client to achieve these capabilities? To answer this question, we make the following technical contributions:
\begin{itemize}
    \item We introduce an adaptive approach called \name that leverages the expertise of different clients on specific data classes throughout the training process. This approach helps address class-imbalance issues by dynamically identifying and harnessing clients' expertise in different phases.
    \item We identify a critical co-learning period (CCP) where we measure the task relevance using the similarity of clients' logits. To filter out irrelevant contributions, we introduce the maximum difference segmentation module (MDS). Additionally, outside the CCP, we employ a scheme based on historical clients' selections to enhance personalization stability.
    \item Through extensive experiments, we demonstrate the superiority of our algorithm, showcasing its effectiveness in personalized federated learning scenarios.
\end{itemize}
\section{Related Work}
In this section, we will present our investigation on existing FL works under non-IID data and various PFL studies. We elaborate their pros and cons, summarizing how they differ from this work. 
\subsection{Data Heterogeneity in Traditional FL}
Data heterogeneity poses a significant challenge in federated learning, even with classical algorithms like FedAvg~\citep{fedavg}, which achieves a global consensus model through local updates and synchronization. Although FedAvg can train models efficiently with distributed data without compromising privacy, the data heterogeneity among clients presents a notable challenge. Previous studies~\citep{li2019noniid} indicate that heterogeneous data not only hampers the model's generalization ability but also hinders the model's convergence.

The performance degradation can be attributed to several factors. Biased distributions, as highlighted in prior research~\citep{learningnot,rethinking}, are interpreted as meaningful information in machine learning. Highly imbalanced data can lead to label bias in classifiers, where decision boundaries are heavily influenced by the majority class. With inconsistent data distributions across clients, decision boundaries also vary, making it challenging to reach a consensus.

To address these issues, traditional FL algorithms have been developed based on FedAvg. FedProx~\citep{FedProx} introduces a proximal term to the local objective function to restrict the updated models to be close to the initial global model. SCAFFOLD~\citep{scaffold} introduces control variables to correct the client-drifts in local updates. MOON~\citep{MOON} takes advantage of the fact that the global model is expected to extract better features by adding a model contrastive-loss term to the loss function to learn a better feature extractor.

However, these algorithms are unable to recognize the similarity of data across different clients, which may introduce irrelevant knowledge. In contrast, our algorithm is able to exploit similarities between clients, thus improving the model's personalization capabilities.

\subsection{Personalized Federated learning}
Personalized federated learning (PFL) allows each client to maintain a private model to enhance personalization capability. Some work adopts the parameter decoupling approach, which treats model as the combination of a feature extractor and a classifier.
To address data heterogeneity, LG-FedAvg~\citep{lgfedavg} only aggregates classifiers, while FedPer~\citep{fedper} and FedRep~\citep{fedrep} only aggregate the feature extractors. Further, based on FedPer, in order to prevent the model from failing to converge, FedVF~\citep{fedvf} allows the classifier to perform low-frequency periodic aggregations, while FedPer++~\citep{fedper++} and FedPAC~\citep{fedpac} employ fine-grained classifier aggregation and feature alignment to mitigate the problem of overfitting. 
Fed-RoD~\citep{fedrod} allows the model to have two classifiers to satisfy both generalization and personalization performance, where the personalized classifier is trained using class-balanced loss and the generic classifier is trained using vanilla loss. FedFA~\citep{fedfa} achieves personalization by fine-tuning classifiers with feature anchors. CReFF~\citep{CReFF} proposes a retraining algorithm to handle the biased classifiers. FedClassAvg~\citep{fedclassavg} only aggregates classifiers to seek a consensus decision boundary, and apply representation learning to stabilize the decision boundary and improve the feature extraction capabilities.

Apart from parameter decoupling, a number of machine learning techniques have also shown great potential in the field of PFL. Meta-learning based FL aims to seek an initial model that can be quickly adapted to local data with few local adaptations. Per-FedAvg~\citep{perfedavg} and pFedMe~\citep{pfedme} focuses on optimizing the personalized models while keeping them close to the global model by introducing a regularization term to the loss function. Knowledge distillation based FL enables efficient communication and heterogeneous models by utilizing the predictive logic of the teacher model to perform local updates. FedDistill~\citep{feddistill} rectifies the local dataset by federated augmentation and performs model synchronization using global-average logits. FedKD~\citep{fedkd} adopts adaptive mutual knowledge distillation with a shared mentor model for model synchronization and exploits  dynamic gradient compression for efficient communication. Prototype-based FL (e.g. FedProto~\citep{fedproto}) collaboratively performs model training by exchanging prototypes instead of gradients.

Unfortunately, the generalization performance of these PFL methods mentioned above will decrease when data is scarce, and only a few of them~\citep{fedper++,fedpac} have considered using classifier collaboration to solve this problem. Compared to them, our algorithm ensures high generalization performance in scenarios with high heterogeneous and sparse data, while also being more lightweight and efficient.
\section{Notations and Preliminary}
In this section, we first provide formal definitions used in PFL training in and then examine the class-imbalance problems based on our theoretical insights on simplified PFL models.
\subsection{Terminology}
Before illustrating our method, we firstly define the symbols that may be used later. Suppose that there are $K$ clients and each client $k$ client holds its own dataset $({\mathbf{x}_i,{y}_i})|_{i=1}^{N_k}\in D_k$. The global dataset $D=\cup_{k=1}^K{D_k}$ has $C$ classes so that every single input is in the Cartesian product of input space $\mathcal{X}$ and label space $\mathcal{Y}$.

Furthermore, we adopt the general practice in computer vision and other domains, by decomposing each client's model parameters ($\mathbf{w}$) into a \textit{feature extractor} ($\bm{\theta}$) and a \textit{classifier} ($\bm{\phi}$); formally, $\mathbf{w}=\{\bm{\theta},\bm{\phi}\}$~\citep{fedper,fedrep,fedrod}.The classifier often indicates the last one or few fully-connected layers, while the rest layers are incorporated into feature extractor. E.g., in convolutional networks, a feature extractor typically consists of series of convolutional layers. We denote $\mathbf{h}$ as the output of feature extractor $f_{\bm{\theta}}:\mathcal{X}\rightarrow\mathcal{H}$. 
%
We adopt another common terminology \textit{proxy}, used in previous works~\citep{fedrs},
to indicate linear weights of different class denoted by $\bm{\phi}=\{\phi_c\}_{c=1}^C$, where each proxy is denoted by $\phi_c$. Besides, we use $z_c$ to denote each \textit{logit} for class $c$, which is a probability. The logits over all classes form the a vector of \textit{logits}, denoted by $\mathbf{z}=(z_1,z_2,\dots,z_C)$, which are computed by $\mathbf{z} = f_\phi:\mathcal{H}\rightarrow\mathbb{R}^{C}$.

We consider that a total of $K$ clients perform the PFL task, where each client $k$ trains its own model by minimizing the local loss $\mathcal{L}_k(\cdot)$. Then the global problem can be formulated as \eqref{eq:PFLobj}:
\begin{eqnarray}
\min_{\mathbf{W}}\mathcal{L}(\mathbf{W}):=\sum_k\frac{\vert D_k\vert}{\vert D\vert}\mathbb{E}_{(\mathbf{x},y)\in D_k}[\mathcal{L}_k(\mathbf{w}_k;(\mathbf{x},y))]
\label{eq:PFLobj}
\end{eqnarray}
where $\mathbf{W}=\{\mathbf{w}_1,\mathbf{w}_2,...,\mathbf{w}_K\}$ is the collection of $k$ model weights, different from training a single model in traditional FL. Equation~\eqref{eq:PFLobj} also naturally leads to the optimal $\mathbf{w}^*$ for each client:
\begin{eqnarray}
\mathbf{w}^*_k=\arg\min_{\mathbf{w}} \mathcal{L}_k(\mathbf{w}_k;(\mathbf{x},y))
\label{eq:localobj}
\end{eqnarray}

For better scalability and interpretability, the local loss $\mathcal{L}_k(\cdot)$ is not explicitly specified here. It can be the cross-entropy loss function used in vanilla FL or it can incorporate unique penalty terms similar to those in knowledge distillation~\cite{fedproto}.

In addition, when describing matrices in the subsequent text, we typically use uppercase bold letters to represent them, while column vectors are represented by lowercase bold letters. The elements of matrices and vectors are denoted using regular letters.

\subsection{Class-Imbalance Training}
\label{subsec:class-imbalance}
{\textbf{Data heterogeneity.}} When discussing data heterogeneity, we always refer to feature skew and label skew. Then the $k$-th client follows $P_k(\mathbf{x},y)=P_k(\mathbf{x}|y)P_k(y)$ which means that the joint distribution of dataset is influenced by both conditional distributions $P_k(\mathbf{x}|y)$ and prior distributions $P_k(y)$. The variation of conditional distribution refers to the different distribution of data features under the same label, while diverse prior distributions across labels and clients reflect the class-imbalance issue in PFL. This also indirectly indicates that methods for managing feature distributions should be compatible with those dealing with label distributions. 

{\textbf{Multi-classification problem under class-imbalance.}} It is important to emphasize that especially in classification problems, the classifier learns the ``patterns'' of features. To better illustrate the impact of class imbalance on model training, for simplicity, let's consider a linear classifier $f_{\mathbf{w}}:\mathcal{X}\rightarrow\mathbb{R}^{C}$. As mentioned in the previous section, $\mathbf{w}$ is actually a matrix composed of different label-specific proxies $\phi_c$. After performing vector multiplication with $\mathbf{x}$, it will be fed into the Softmax function to output logits:
\begin{align}
\mathbf{z}   &=softmax(\mathbf{w}^\top\mathbf{x}) \notag\\
    &=(z_1,z_2,\dots,z_C) \notag\\
    &=\left(\frac{e^{\phi_1^\top\mathbf{x}}}{\sum e^{\phi_c^\top\mathbf{x}}},\frac{e^{\phi_2^\top\mathbf{x}}}{\sum e^{\phi_c^\top\mathbf{x}}},\dots,\frac{e^{\phi_C^\top\mathbf{x}}}{\sum e^{\phi_c^\top\mathbf{x}}}\right)
\label{eq:softmax}
\end{align}

When the loss function is cross-entropy, given \eqref{eq:softmax}, the partial derivatives with respect to each proxy $\phi_c$ can be derived, which will vary depending on the certain class of training samples, thus affecting the backpropagation process:
\begin{align}
\phi_c' &=\phi_c-\eta\frac{\partial\mathcal{L}}{\partial\phi_c}  =\begin{cases}
            \phi_c+\eta(1-z_{c_0})\mathbf{x},&c_0=c \\
            \phi_c-\eta z_{c_0}\mathbf{x},&c_0\neq c \\
        \end{cases}
\label{eq:CEloss}
\end{align}


This means that when the data distribution is extremely skewed, such as in a ten-class classification problem where only data with a single label $c$ is used for training, the update of the proxy corresponding to $c$ follows the first case of equation~(\ref{eq:CEloss}). This leads to a significant amount of training for $\phi_c$, aligning the proxy with the features of class $c$ and gradually learning the ``pattern'' of class $c$, resulting in a tendency towards high prediction values. On the other hand, the update process for the remaining proxies follows the second case, gradually moving away from the ``pattern'' of class $c$, and failing to learn their own ``patterns''. 

{\textbf{Prediction bias.}} We then examine the prediction bias under class-imbalanced circumstances through the concept of prototype~\citep{fedproto,fedfa}. Suppose that for any random data sample $\tilde{\mathbf{x}}_c\in\mathbb{R}^d$ belonging to a certain class $c\in[C]~(d>C)$, there exists a prototype $\bar{\mathbf{x}}_c=\mathbb{E}_{\tilde{\mathbf{x}}_c\sim P_k(\mathbf{x}|c)}[\tilde{\mathbf{x}}_c]$. If the prototypes formed by different categories are linearly dependent on each other, 
%
the classifier will inevitably fail to distinguish between these two distinct concepts $c_1$, $c_2$ from random data $\tilde{\mathbf{x}}_{c_1}$, $\tilde{\mathbf{x}}_{c_2}$. In such cases, the dimension of the input needs to be to expand to incorporate additional information that aids in discrimination. In the other case, where the prototypes are orthogonal to each other, the rank of the matrix formed by prototypes from different categories will be equal to the number of classes $C$, and the prototypes will be linearly independent from each other. This is analogous to performing 0-1 classification where the data points have no correlation. In this scenario, even if only one sample is provided for each label, the classifier can perform excellently. 

However, in reality, prototypes can be linearly dependent in classification tasks. If training is conducted in the presence of class-imbalance, linear classifiers continuously fit proxies to their respective prototypes, assigning higher weights when predicting their own data. Due to the presence of linear relationships among prototypes and unequal learning of different proxies by (\ref{eq:CEloss}), it is highly likely to confuse the classes corresponding to two prototypes that have a linearly dependent relationship. This is where prediction bias arises: when clients make predictions, they often give their dominant classes larger prediction values. Conversely, {\em two classifiers with the same prediction bias are likely to share similar dominant classes}. 

Given these preliminary understanding of potential results due to class-imbalance, we are ready to propose our algorithm in Section~\ref{subsec:relevantmatching} and \ref{subsec:CCP}, where the above analysis will be further validated.
\section{Our Method: FedReMa}
Due to the class-imbalance issue in PFL training, aggregating the classifiers of clients that have very diverse expertise across classes might be detrimental. Based on the analysis in Section~\ref{subsec:class-imbalance}, a group of clients with similar biases may share a set of dominant classes in common. It is then feasible to assess their data distributions through the similarity of logits. By this means, we can efficiently identify the most relevant peers for each client and establish their training dependencies. Performing classifier collaboration based on these inter-client dependencies can potentially reduce the overfitting of proxies. Our design goal is to enhance each model's personalizability on dominant classes, preserving their preferences tailored to local datasets, while improving the generalization performance through the peers' knowledge on the client's non-dominant classes. 

\begin{figure*}[t]
\centering
\includegraphics[width=14cm]{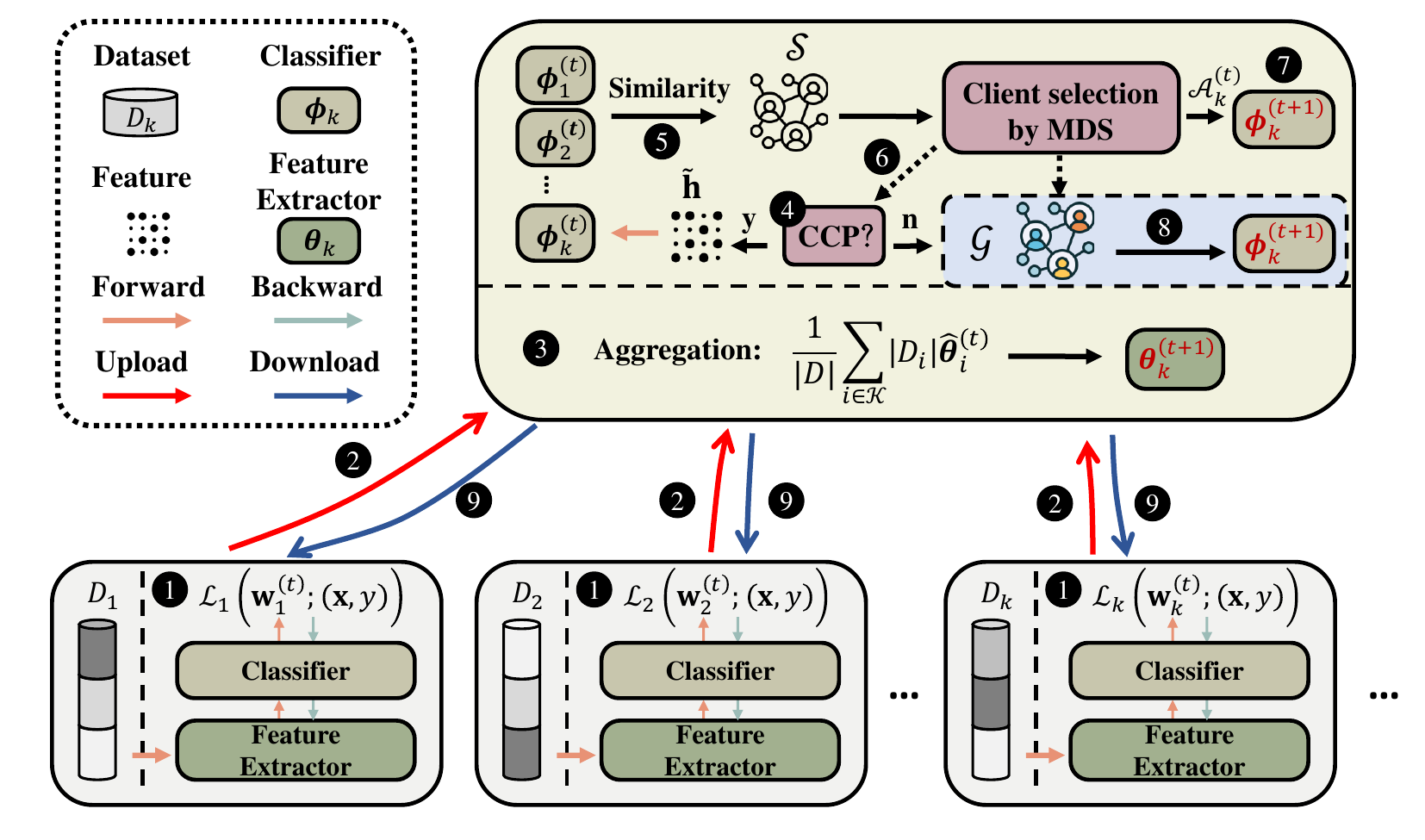}
\caption{Workflow of \name. (1)~Local training. (2)~Upload model to server. (3)~Aggregate global feature extractors. (4)~Determine whether the current period is a critical learning period, if so, go to (5), otherwise go to (8). (5)~Measure the task relevance $\mathcal{S}$ using similarity of their logits, then perform MDS to filter out irrelevant clients. At the same time, the server will record the client's historical dependencies $\mathcal{G}$. (6)~Update \textit{CCP} status. (7)~Aggregate personalized classifiers by $\mathcal{A}_k^{(t)}$. (8)~Aggregate personalized classifiers based on historical dependencies $\mathcal{G}$. (9)~Server communicates personalized models to all clients.}
\label{fig:workflow}
\end{figure*}

To realize this, we propose \name, an adaptive method that leverages a novel relevant matching algorithm (named as MDS shown in Algorithm \ref{algo:MDS}) for classifier synergy and carefully identified critical co-learning period (CCP), adapting to the dynamics of clients' prediction expertise; the feature extractor aggregation is the same as standard FedAvg. As shown in Figure~\ref{fig:workflow}, after each local training is completed, the server judges whether the current round is the critical co-learning period (Section \ref{subsec:CCP}) and adopts MDS if it is the case; otherwise, a distinct classifier synergy algorithm leveraging historical peer selections is performed (Section \ref{subsec:counting}). 

\subsection{Relevant Matching}
\label{subsec:relevantmatching}
{\textbf{Local Training.}} As shown in lines 1-6 of Algorithm~\ref{algo:FedReMa}, like most federated learning algorithms, we assume that during communication round $t\in[1,T]$, the clients will perform multiple epochs of parameter updates. 
At the beginning of $t$, the clients receive a personalized model from the server and initialize it as $\mathbf{w}_k^{(t)}\leftarrow\mathbf{w}_k^{(t-1)}$. They then conduct $E$ epochs of local training and transmit the updated model $\hat{\mathbf{w}}_k^{(t)}=\{\hat{\bm{\theta}}_k^{(t)}, \hat{\bm{\phi}}_k^{(t)}\}$ back to the server for the next step of the algorithm. Here, we have the option to employ simple stochastic gradient descent (SDG) for training, or explore different training strategies. The distinctions between various approaches will be observed in Section~\ref{sec:Performance}.

{\textbf{Relevant Matching.}} In this section, we will introduce how to perform the relevant matching algorithm using the logits similarity of the uploaded classifier set $\{\bm{\phi}_k^{(t)}\}_{k\in\mathcal{K}}$. For the sake of brevity, we will ignore the superscript ${(t)}$ of some variables. 

Based on the analysis from Section~\ref{subsec:class-imbalance}, due to training under class-imbalance, there is an inequality in the sufficiency of training for proxies corresponding to different classes. This leads to the models in the early stages being prone to prediction bias. Even when we input a randomly generated feature, different classifiers will provide biased logits that are strongly correlated with the local class distribution. 

We test each client's classifier response by inputting a feature $\tilde{\mathbf{h}}$ where each element is independently sampled from a uniform distribution in the range [0,1). We then construct a soft prediction method, using $\mathbf{p}_{k}=softmax(\mathbf{\phi}^\top_k\mathbf{h}/M)$ to control the sharpness of logits via a temperature parameter $M$ (Unless otherwise specified, we set $M=0.5$). We establish this by borrowing the idea from the concept of response-based knowledge in knowledge distillation~\citep{gou2021knowledge}.


The new $\mathbf{p}_{k}$ we call it {\it soft logits}. Consider a connected graph $\mathcal{S}$ that can be fully represented by a relativity matrix $\mathbf{S}\in\mathbb{R}^{K\times K}$, where the weights are obtained through the computation of cosine similarity between the soft logits of different clients:
\begin{equation}
s_{k_1,k_2}=\frac{\mathbf{p}_{k_1}\cdot\mathbf{p}_{k_2}}{\Vert \mathbf{p}_{k_1}\Vert\Vert \mathbf{p}_{k_2}\Vert}
\label{eq:simlarity}
\end{equation} 

Due to the properties of the softmax function, the similarity values calculated in the end are in the range of [0,1]. For each client $k$, there exists a relevance vector $\mathbf{s}_k=(s_{k,1},\dots,s_{k,K})$. We employ the {\em maximum difference segmentation} method (MDS, shown as Algorithm~\ref{algo:MDS}) to obtain the set of most relevant clients $\mathcal{A}_k$ for client $k$, using the maximum difference as the boundary for client selection.

\begin{algorithm}[t]
    \caption{Maximum difference segmentation (\texttt{MDS})}
    \label{algo:MDS}
    \LinesNumbered
    \KwIn{Relativity vector $\mathbf{s}=(s_{1},\dots,s_{K})$}
    \KwOut{Set of the most relevant nodes $\mathcal{A}$ and maximum difference $\Delta s^*$}
    Sort the relativity element:
    $\mathbf{idx}=\text{SortIndices}(\mathbf{s})=(\text{idx}_1,\dots,\text{idx}_K)$,
    $\text{Sort}(\mathbf{s})=(s_{\text{idx}_1},\dots,s_{\text{idx}_K})$;\\
    Calculate difference
    $\Delta s_{\text{idx}_i}=s_{\text{idx}_{i+1}}-s_{\text{idx}_i}$;\\
    Get ${m}=\arg\max_{i\in[K-1]}{\Delta s_{\text{idx}_{i}}}$
    $\Delta s^* = \Delta s_{\text{idx}_{m}}$;\\
    \textbf{Return} $\mathcal{A}=\{\text{idx}_{m},\text{idx}_{m+1},\dots,\text{idx}_{K}\}$ and $\Delta s^*$;
\end{algorithm}

Then, the server performs distinct aggregation methods on the uploaded feature extractors and classifiers:
\begin{align}
    \bm{\theta}^{(t)}&\leftarrow\frac{1}{\vert D\vert}\sum_{i\in\mathcal{K}}\vert D_i\vert\bm{\hat{\theta}}_{i}^{(t)}\label{eq:body}\\
    \bm{\phi}^{(t)}_k&\leftarrow\frac{1}{\vert D_{\mathcal{A}^{(t)}_k}\vert}\sum_{i\in\mathcal{A}^{(t)}_k}\vert D_i\vert\bm{\hat{\phi}}_{i}^{(t)}\label{eq:head1}
\end{align}

The server aggregates a robust global feature extractor in equation~(\ref{eq:body}), while equation~(\ref{eq:head1}) ensures the personalized aggregation of classifiers. 


\begin{figure*}[htp]
    \centering
    \subfloat[$t=1$]{\label{fig:round1}
        \centering
        \includegraphics[width=0.24\linewidth]{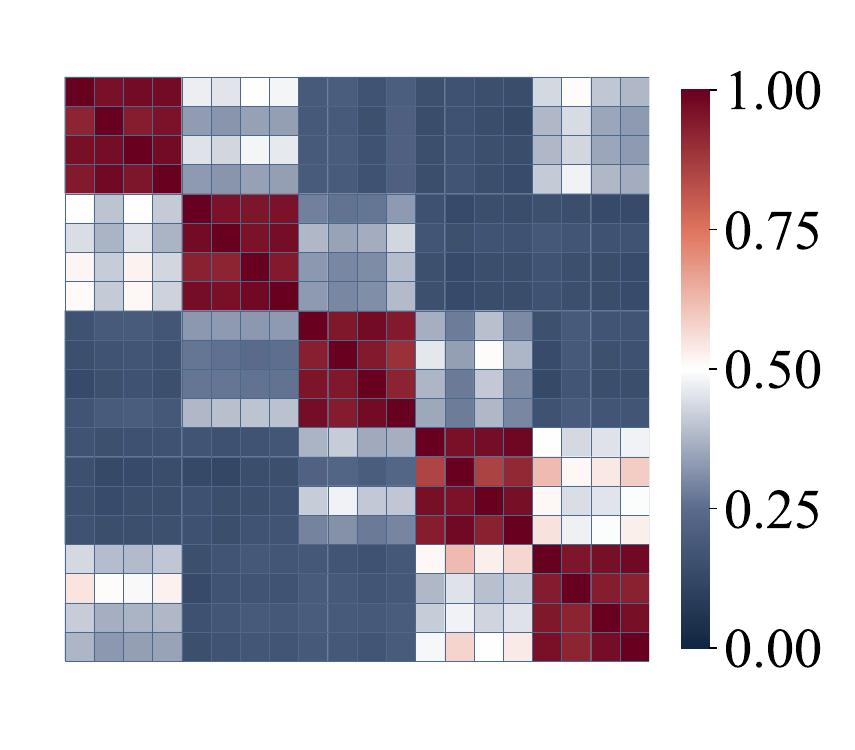}
    }
    \hfill
    \subfloat[$t=50$]{\label{fig:round50}
        \centering
        \includegraphics[width=0.24\linewidth]{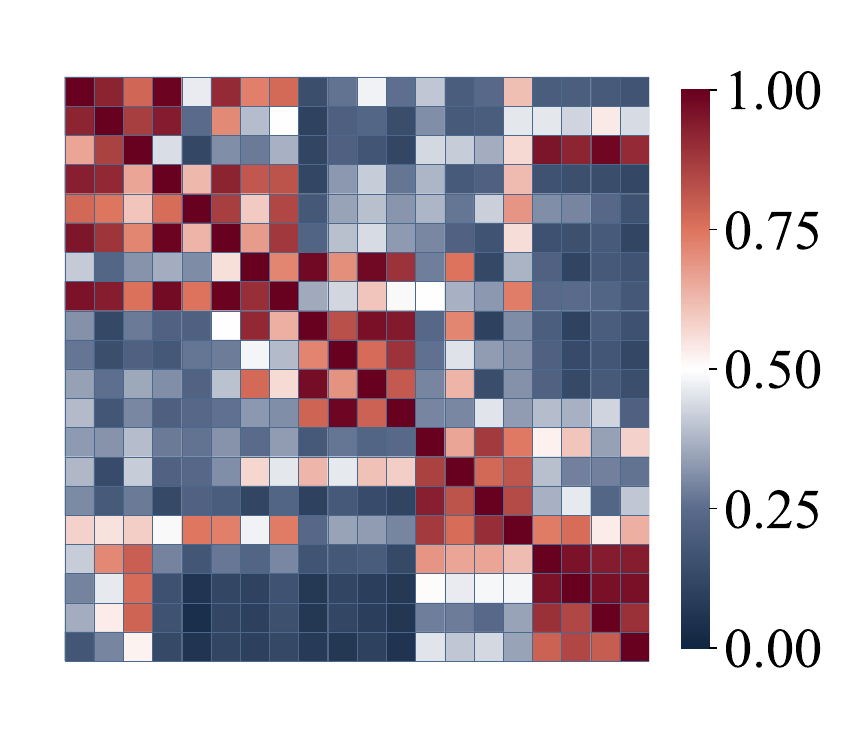}
    }
    \hfill
    \subfloat[$t=100$]{\label{fig:round100}
        \centering
        \includegraphics[width=0.24\linewidth]{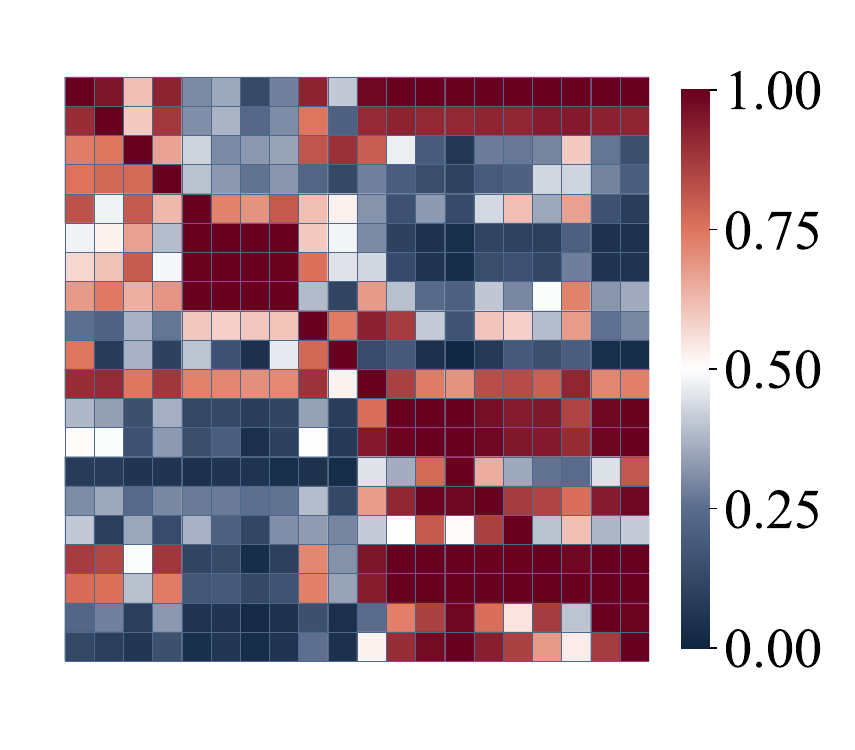}
    }
    \hfill
    \subfloat[$t=150$]{\label{fig:round150}
        \centering
        \includegraphics[width=0.24\linewidth]{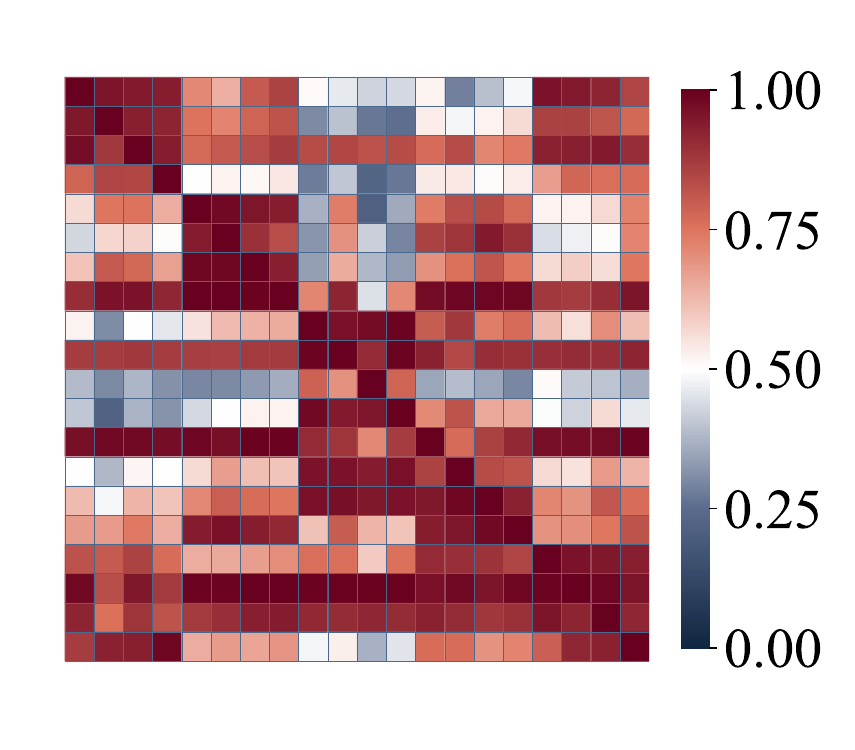}
    }
    \caption{The heatmap of $\mathbf{S}^{(t)}$ on different communication round $t$ (FedAvg)}
    \label{fig:heatmap}
    \vspace{10pt}
\end{figure*}

\subsection{{Critical Co-learning Period}} 
\label{subsec:CCP}
As shown in Figure~\ref{fig:heatmap}, we gain another key finding through experiments: {\em As the training progresses, the effectiveness of relevant matching degrades}. This can imply that models in the later stages of training already have the ability to identify non-dominant classes. The model performance might be undermined we continuously adopt the previous metric and method for classifier synergy. Therefore, we propose a critical co-learning period (CCP) method based on the historical maximum similarity difference given by MDS. We define the average maximum similarity difference as the average value of $\{\Delta s^{t}_k\}_{k\in\mathcal{K}}$ for all clients in round $t$.

\begin{equation}
    \Delta \bar{s}^t=\sum_{k\in\mathcal{K}}\Delta s^{t}_k
    \label{eq:AMSD}
\end{equation}
The critical co-learning period can be determined via~(\ref{eq:CCP}):
\begin{align}
CCP &=\begin{cases}
            True,& \frac{\Delta\bar{s}^{t}}{\max_{\tau\in[t]}\Delta\bar{s}^{\tau}}>\delta\\
            False,&\text{otherwise} \\
        \end{cases}
\label{eq:CCP}
\end{align}
Here, $\delta$ is a constant threshold. It is evident that the determination of \textit{CCP} is triggered by the decrease of $\Delta \bar{s}^t$. Once we are unable to differentiate relevant clients based on similarities, the co-learning in this stage becomes ineffective. If the MDS algorithm continues to be applied, there will be a degradation in accuracy. Another way to effectively aggregate the classifiers must be used.


\subsection{Personalization via Historical Peer Matching}\label{subsec:counting}
The primary purpose of setting the critical co-learning period is (1) to simplify the computational cost in the later phase, and (2) to maintain personalization. During the \textit{CCP}, server maintains a dependency map $\mathcal{G} \in \mathbb{R}^{K \times K}$, where each column vector $\bm{g}_k=(g_{k,1},\dots,g_{k,K})^\top$ represents the number of times client $k$ selects other clients as its relevant peers. When client $j$ is repeatedly selected by client $i$, a high value of $g_{i,j}$ will indicate a strong dependency relationship of $i$ on $j$.

When \textit{CCP} is determined to be $False$, we adopt an algorithm that leverages the historical matching decisions. The rationale is that: We can only utilize the MDS algorithm for client selection to facilitate classifier collaboration during \textit{CCP}. However, after \textit{CCP}, the clients may have accumulated a certain level of knowledge, causing the distinction between soft logits of relevant clients to become ambiguous. 
Thus, we $\mathcal{G}$ is a historical dependency map that utilizes the $\mathcal{A}_k^{(t)}$ when the most relevant clients is still distinguishable. It records the number of times a particular client selects others, which itself serves as a form of weight information.

\begin{equation}
    \bm{\phi}^{(t)}_k\leftarrow\frac{1}{\sum_{i\in\mathcal{K}}g_{k,i}}\sum_{i\in\mathcal{K}}g_{k,i}\cdot\bm{\hat{\phi}}_{i}^{(t)}
    \label{eq:head2}
\end{equation}

With the help of dependency map $\mathcal{G}$, we can perform classifier collaboration directly via Equation~(\ref{eq:head2}). In the experiments presented later in the Section~\ref{sec:Performance}, we can observe that this design effectively reduces computational complexity and training time.

\begin{algorithm}[t]
	\caption{PFL via the Most Relevant Peers (\name)}
        \label{algo:FedReMa}
	\LinesNumbered 
	\KwIn{initial model $\mathbf{w}=\{\bm{\theta},\bm{\phi}\}$, client number $K$, global round $T$, local epoch $E$}
	\KwOut{Personalized model $\mathbf{w}_k^*$} 
        \For{$t=1$~\textbf{\textup{to}}~$T$}{
            \ForEach{client $k\in\mathcal{K}$ ~\textbf{\textup{in parallel}}}{
                Initialize the local model $\mathbf{w}_k^{(t)}\leftarrow\mathbf{w}_k^{(t-1)}$;\\
                $\hat{\mathbf{w}}_k^{(t)}\leftarrow$\texttt{ClientTrain}$(\mathbf{w}_k;D_k)$;\\
                Communicaticate $\hat{\mathbf{w}}_k^{(t)}$ back to the server;\\
    	  }
            Server calculates $\mathbf{S}^{(t)}$;\\
            Server aggregates the feature extractor by Eq.~(\ref{eq:body});\\ 
            \eIf{\textit{CCP}}{
                \For{each client $k\in\mathcal{K}$}{
                    Get relevant set $\mathcal{A}^{(t)}_k\leftarrow${\texttt{MDS}}$(\mathbf{s}_k^{(t)})$;\\
                    Personalization by Eq.~(\ref{eq:head1});\\
                }
                Server counts $\mathcal{G}$;\\
                Check \textit{CCP} by Eq.~(\ref{eq:CCP});\\
            }{
                Personalization by Eq.~(\ref{eq:head2});\\
            }
            Server communicates $\mathbf{w}_k^{(t)}=\left\{\bm{\theta}^{(t)},\bm{\phi}_k^{(t)}\right\}$ to all clients;
        }

\end{algorithm}
\section{Experiment}
In this section we will present our experimental results.

\subsection{Settings}

\textbf{Dataset.}
We evaluate the algorithm using three classic computer vision classification datasets: MNIST, FMNIST, and CIFAR-10. These datasets consist of 50000 training images (60000 For FMNIST) with 10 classes. 
To control data heterogeneity across local datasets, we adopt the same data partitioning scheme as \citep{scaffold,fedpac}.
Specifically, while ensuring dataset equality, we divide all the clients into five groups. In each group, three labels are designated as dominant labels, with a proportion parameter $s\%$ controlling the fraction of IID data, while the remaining $(1-s)\%$ data belong to dominant labels.

\textbf{Baselines.}
We compare the following methods with our \name: (1)~Local (no model aggregation). (2)~FedAvg~\citep{fedavg}, (3)~FedProx~\citep{FedProx}, (4)~MOON~\citep{MOON}, (5)~FedProto~\citep{fedproto}, (6)~FedPer~\citep{fedper}, (7)~Per-FedAvg~\citep{perfedavg}, (8)~FedBABU~\citep{fedbabu},  (9)~FedPAC~\citep{fedpac}, (10)~FedDistill~\citep{feddistill}.

\textbf{Model Settings.}
For MNIST and FMNIST, we use a CNN model consisting of 2 convolutional layers and 2 fully connected layers. For CIFAR-10, we use a simplified AlexNet for training. Specifically, we have reduced the size of the convolutional kernels and omitted the use of Dropout and Local Response Normalization (LRN) layers. This version of AlexNet is better suited for CIFAR-10. We consider the fully connected layers of each network as classifiers, and the remaining convolutional layers as feature extractors.

\textbf{Hyperparameters.}
We use mini-batch stochastic gradient descent (SGD) as the optimizer for all methods. The number of communication rounds is set to $T=500$, and the number of local training epochs between consecutive communications are set to $E=5$. The batch size is set to 100, and the learning rate is set to $\eta=0.01$.
For FedProx, $\mu$ is set to 1. For MOON, $\mu$ is set to 1 (5 for the CIFAR-10). The regularization $\lambda$ is 1 for both FedProto and FedPAC. For other hyperparameters not mentioned, we adopt the best of them reported in their respective original papers.

\textbf{Evaluation Metrics.}
In contrast to conventional FL where the global is finally tested on a centralized dataset,
PFL measures the personalized accuracy over clients under individual local datasets. 
All experiments were conducted on a lab server with a NVIDIA GeForce RTX 3090 GPU. 

\subsection{Performance Evaluation}
\label{sec:Performance}
\textbf{Comparison across different datasets.} To simulate a highly heterogeneous environment, we initially set the sample size for each client to be 600 and divided it via a 4:1 ratio for both training and testing. We conducted experiments with $s=0.2$. Each set of experiments were repeated 5 times with different random seeds. 
Table~\ref{table:all} shows the performance comparison. Our \name achieves the highest accuracy in the majority of settings. It is worth noting that FedAvg remains a strong baseline, exhibiting good performance even in scenarios with high heterogeneity and limited samples.

\begin{table}
    \caption{Accuracy on different datasets.}
    \centering
    \begin{tabular}{rrccc} %
        \toprule[1pt] 
        
        &\#\tbf{Method}&\tbf{MNIST} &\tbf{FMNIST} &\tbf{CIFAR-10}          \\
        \midrule
        \multirow{4}{*}{\tbf{TFL}}
        &Local       &94.3±0.2\% &86.0±0.2\% &58.9±0.1\%  \\
        &FedAvg      &98.4±0.1\% &86.1±0.3\% &61.9±0.6\%  \\
        &FedProx     &97.8±0.2\% &84.8±0.1\% &61.6±0.8\%  \\
        &MOON        &98.4±0.2\% &86.0±0.2\% &61.9±0.7\%  \\
        \cmidrule(lr){1-5}
        \multirow{6}{*}{\tbf{PFL}}
        &FedProto    &95.6±0.1\% &80.8±0.4\% &56.1±0.4\%  \\
        &FedPer      &94.8±0.1\% &86.1±0.1\% &59.5±0.6\%  \\
        &Per-FedAvg  &97.6±0.3\% &88.1±0.3\% &62.3±2.1\%  \\
        &FedDistill  &96.3±0.1\% &87.0±0.1\% &61.2±0.5\%  \\
        &FedBABU     &97.1±0.2\% &83.1±0.2\% &53.6±0.8\%  \\
        &FedPAC      &98.4±0.1\% &86.9±0.4\% &64.9±0.1\%  \\
        \cmidrule(lr){1-5}
        \multirow{1}{*}{\tbf{Ours}}
        &FedReMa      &\bf{98.5±0.1}\% &\bf{88.2±0.3}\% &\bf{65.4±0.3}\%  \\
        \bottomrule[1pt]
    \end{tabular}
    \label{table:all}
\end{table}

\textbf{Comparison of varying data heterogeneities.} 
We further conducted experiments varying data heterogeneity, shown in Table~\ref{table:heterogeneity}. Each client had only 100 samples, imposing limitations on model generalization.

\begin{figure}[t]
    \centering
    \subfloat[$s=0.2$]{\label{fig:trains02}
        \centering
        \includegraphics[width=0.48\linewidth]{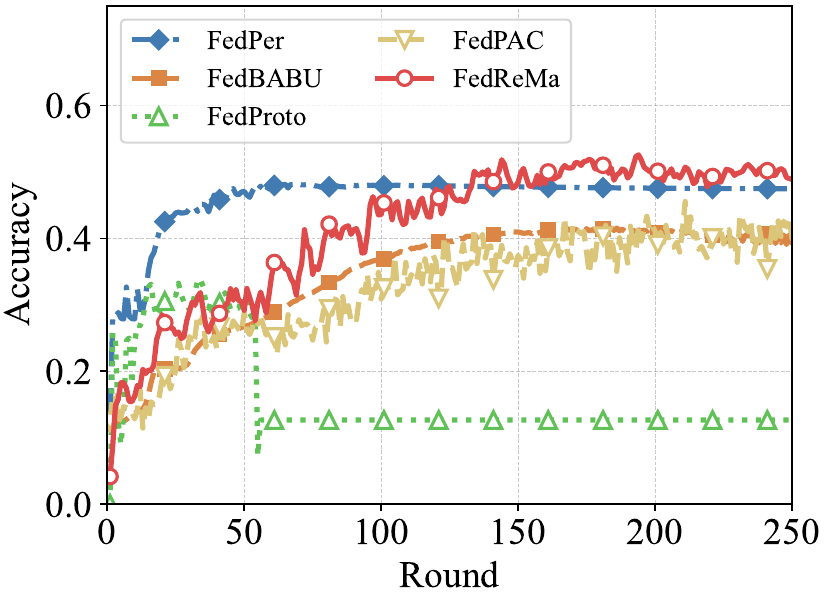}
    }
    \hfill
    \subfloat[$s=0.4$]{\label{fig:trains04}
        \centering
        \includegraphics[width=0.48\linewidth]{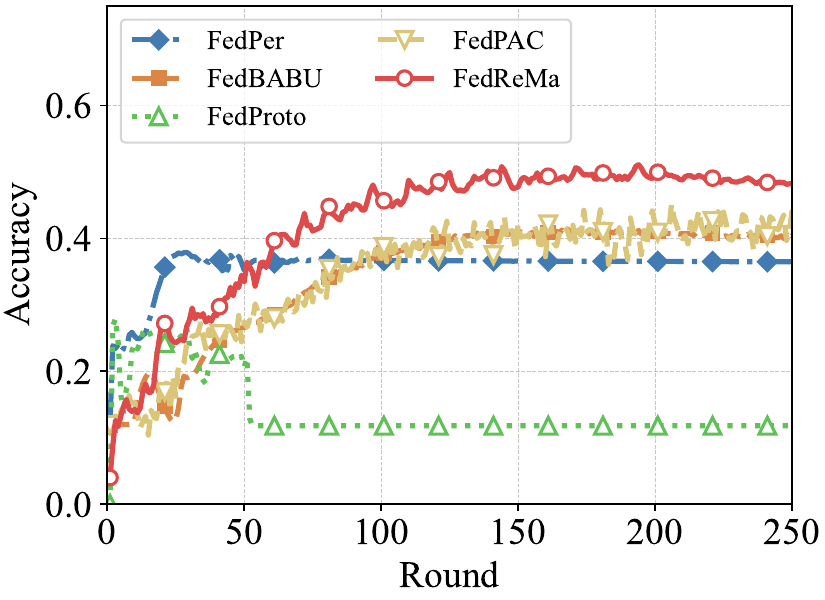}
    }
    \hfill
    \subfloat[$s=0.6$]{\label{fig:trains06}
        \centering
        \includegraphics[width=0.48\linewidth]{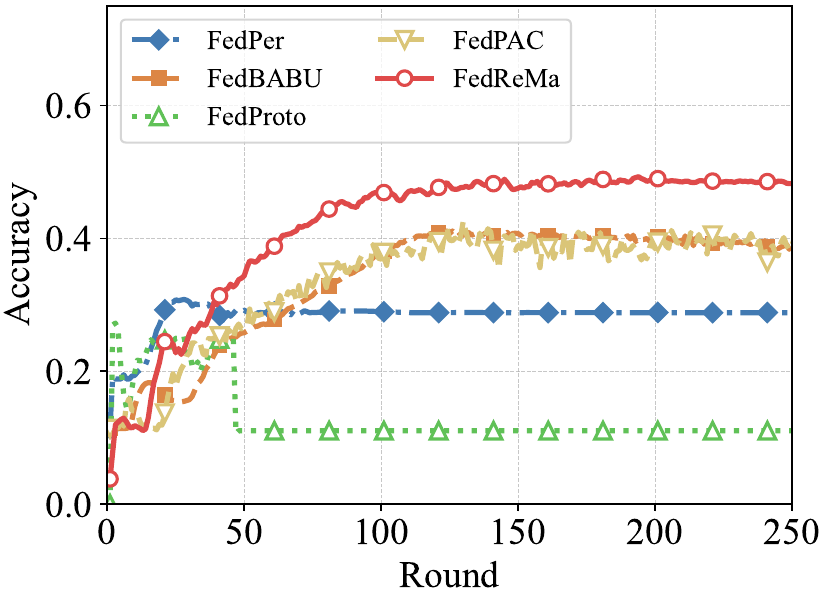}
    }
    \hfill
    \subfloat[$s=0.8$]{\label{fig:trains08}
        \centering
        \includegraphics[width=0.48\linewidth]{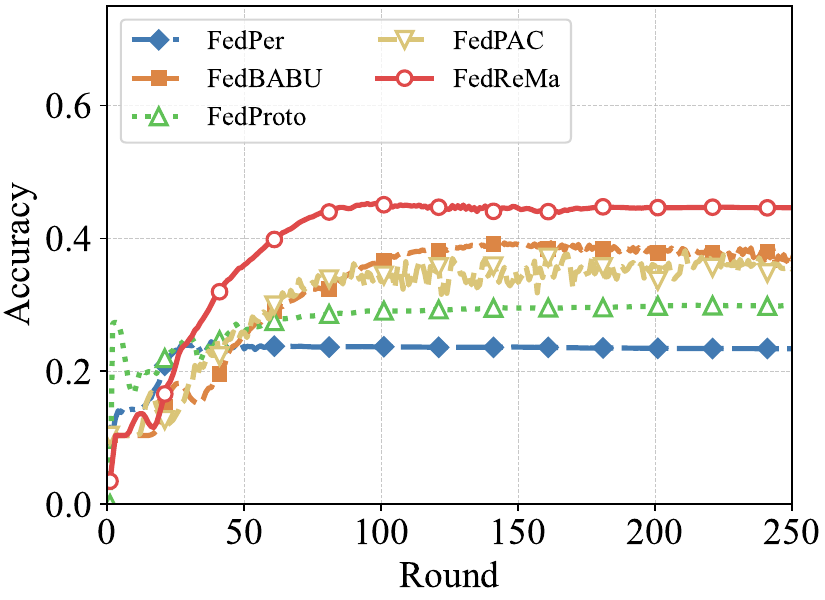}
    }
    \caption{The test accuracy while training process.}
    \label{fig:trains}
    \vspace{22pt}
\end{figure}

\name still maintains the best performance when different values of $s$ are chosen. 
Note that the generalization performance of the algorithms is greatly challenged in the case of sparse data. When the data heterogeneity is high ($s=0.2$), various methods can still maintain a certain level of accuracy. FedPer and FedPAC perform well, while FedProto's accuracy is significantly affected, even experiencing a complete collapse in the later stages of training (Figure~\ref{fig:trains02}, \ref{fig:trains04} and \ref{fig:trains06}). We believe this is due to the inability to achieve unbiased estimation of prototypes with sparse data. As the data homogeneity increases, FL methods stand out while PFL methods experience varying degrees of accuracy degradation.

However, when the data approaches homogeneous, \name also experiences performance degradation. It is reasonable because we do not introduce additional techniques apart from adaptive personalized aggregation of classifiers, for reducing time complexity. Therefore, under less heterogeneous data, the \textit{CCP} ends prematurely, resulting in insufficient personalization. 

\begin{table}
    \caption{Accuracy on CIFAR-10 with different data heterogeneity.}
    \centering
    \begin{tabular}{rcccc}
        \toprule[1pt]         
        \#\tbf{Method} &\tbf{s=0.2} &\tbf{s=0.4} &\tbf{s=0.6} &\tbf{s=0.8} \\
        \midrule
        Local       &48.1±0.8\% &38.1±0.6\% &30.2±0.4\% &24.2±0.5\% \\
        FedAvg      &46.4±0.9\% &47.1±1.2\% &47.4±1.2\% &45.4±1.3\% \\
        FedProx     &40.7±0.6\% &41.9±0.5\% &41.6±0.8\% &41.3±0.4\% \\
        MOON        &46.4±0.8\% &47.1±1.2\% &47.4±1.1\% &45.5±1.2\% \\
        \cmidrule(lr){1-5}
        FedProto    &36.7±3.0\% &29.2±0.8\% &30.9±2.7\% &30.6±0.9\% \\
        FedPer      &47.6±0.8\% &37.8±0.9\% &30.9±0.9\% &25.1±0.6\%\\
        FedBABU     &41.3±0.5\% &41.1±0.5\% &40.9±0.6\% &40.1±0.7\%\\
        FedPAC      &48.0±2.1\% &45.7±1.2\% &41.6±0.8\% &38.3±0.3\%\\
        \cmidrule(lr){1-5}
        FedReMa     &\bf{52.9±1.2}\% &\bf{50.2±0.7}\% &\bf{47.6±1.4}\% &\bf{46.5±1.8}\%\\
        \bottomrule[1pt]
    \end{tabular}
    \label{table:heterogeneity}
\end{table}

\textbf{Sensitivity.}
We also conducted experiments varying $\delta$, our algorithm's key hyperparameter, which controls the epochs of the CCP. 
The experimental results are shown in the Table~\ref{table:delta}. ``Final accuracy'' refers to the average accuracy of the last 5 communication rounds.
\begin{table}
    \caption{Sensitivity of $\delta$.}
    \centering
    \begin{tabular}{llllll}
        \toprule[1pt] 
        $\delta$       &0.1     &0.3     &0.5     &0.7     &0.9     \\
        \midrule
        Best Acc        &63.36\%  &64.13\%  &\bf{65.40}\%  &65.26\%  &65.06\%    \\
        Final Acc       &61.64\%  &60.73\%  &\bf{63.77}\%  &62.30\%  &63.03\%    \\
        Epochs of CCP    &500      &132      &36       &16       &7          \\
        \bottomrule[1pt]
    \end{tabular}
    \label{table:delta}
\end{table}


The best accuracy of \name doesn't vary significantly when $\delta$ increases, as the best model occurs during the early phase when the MDS algorithm is still being executed. 
%
When $\delta=0.1$, the running time for \name is 1403.08s. While for $\delta=0.9$, it is 760.89s (54.23\%). Maintaining personalization can be observed from the final accuracy decrease. An unreasonably small $\delta$ can lead to degradation to FedAvg, while a too large $\delta$ may fail to provide a good enough $\mathcal{G}$ in the second stage. Overall, although our algorithm is relatively robust to the value of $\delta$, $\delta=0.5$ is the best choice in this experimental configuration.

\textbf{Compatibility with FedPAC.} The previous experiments demonstrated the excellent performance of \name with the help of MDS algorithm and critical co-learning period. However, \name itself is a server-side algorithm and does not introduce any new methods on the client side. Thus we have conducted additional testing on the FedPAC method, which has also undergone classifier collaboration. We can confirm that if we combine the techniques in FedPAC with our \name, the performance can be further enhanced. 
\begin{table}
    \caption{Accuracy and Runtime Comparison with FedPAC and extended versions of our \name.}
    \centering
    \begin{tabular}{llll}
        \toprule[1pt]
        \#\tbf{Method}    &\tbf{Accuracy} &\tbf{Runtime}&      \\
        \midrule
        FedReMa   &65.40±0.32\% & 796.5s \\
        FedPAC    &64.93±0.13\% & 4464.3s \\
        FedReMa+F        &66.31±0.55\% & >3000s\\
        FedReMa+A\&F     &69.73±0.98\% & >3000s\\
        \bottomrule[1pt]
    \end{tabular}
    \label{table:extended}
\end{table}

The training strategies on client-side in FedPAC are as follows:

\begin{enumerate}
\item \textbf{Feature alignment}: 
As shown in (\ref{eq:FA}), an additional feature alignment penalty term is added in FedPAC.
\begin{align}
    \mathbf{w}_k^*  &=\arg\min_\mathbf{w}\mathcal{L}_k(\mathbf{w}_k;(\mathbf{x},y)) \notag\\
                    &=\arg\min_\mathbf{w}\mathcal{L}_{CE}(\mathbf{w}_k;(\mathbf{x},y))+\lambda\mathcal{L}_d(\bar{h};h)
                    \label{eq:FA}
\end{align}
where $\mathcal{L}_d$ is usually the L2 norm. This approach requires additional communication and feature aggregation on the server side.

\item \textbf{Alternately train feature extractors and classifiers}: One advantage of alternating optimization (optimizing the classifier first, then the feature extractor) is that each optimization step of the classifier is performed in the same feature space. Thus, comparing the logits of the classifier becomes more meaningful.
\end{enumerate}

It is not difficult to see that these are all strategies that are beneficial for normalizing features. By incorporating these client training strategies, we aim to enhance the performance of \name and showcase the potential for even better results. The result is shown in Table~\ref{table:extended}, where "F" means {\em Feature Alignment} and "A" means {\em Alternately Training}. Those strategies indeed lead to better performance. Compared with FedPAC, our method has a huge advantage in computational complexity.

\textbf{Ablation study.}
We finally verify the effectiveness of each component of our algorithm. As shown in the Table~\ref{table:ablation}, the two components we designed complement each other and both achieve satisfactory accuracy. This means that both components are well suited to the task of parameter personalization.
\begin{table}
    \caption{Ablation experiment.}
    \centering
    \begin{tabular}{llllll} %
        \toprule[1pt] 
        
        \#\tbf{Method} &\tbf{Best Acc} &\tbf{Final Acc} &\tbf{Runtime}           \\
        \midrule
        None(FedAvg) &61.95±0.65\% &61.80\% &691.58s \\
        +MDS         &63.36±0.79\% &61.64\% &1403.08s\\
        +MDS\&CCP    &65.40±0.32\% &63.77\% &796.54s\\
        \bottomrule[1pt]
    \end{tabular}
    \label{table:ablation}
\end{table}

\section{Conclusion}
In this paper, we proposed \name, an improved personalized federated learning framework by strategically leveraging the expertise of different clients on specific data classes throughout the training process. \name addresses the issue of class-imbalance by dynamically identifying and harnessing clients’ expertise in different phases. By performing a novel method that combines relevant matching among clients using similarities of clients' logits and the judgement of critical co-learning period, the server can efficiently identify clients with similar tasks and enhance personalized accuracy through the personalized aggregation of classifiers. \name outperforms existing methods significantly across various personalized federated learning scenarios. 

There are several possible future research directions. For example, extending the algorithm to more general tasks, providing stronger theoretical analysis for the algorithm, etc. Another example is to explore how the correlation between features of different categories will affect the update of the classifier, and maybe there is a possible way to optimize the aggregation of the classifier by linking the correlation between features. Furthermore, there are many interesting potential optimization directions to extend our algorithm. For example, we can employ more fine-grained aggregation of different proxies of the classifier and layer-by-layer personalization based on feature similarities across clients.
\begin{ack}
This work was supported in part by Guangdong Basic and Applied Basic Research Foundation (2024A1515010161, 2023A1515012982, 2023B1515120058, 2021B151520008), Guangzhou Basic and Applied Basic Research Program (2024A04J6367), NSFC grant (62102460), Young Outstanding Award under the Zhujiang Talent Plan of Guangdong Province, and the Ministry of Education, Singapore, under its Academic Research Fund (AcRF RG91/22) and NTU Startup. 

All experiments were conducted at RTAI cluster in School of Computer Science and Engineering, Sun Yat-sen University.
\end{ack}




\bibliography{mybibfile}

\end{document}